\documentclass{article}
\usepackage{arxiv}

\usepackage[utf8]{inputenc} 
\usepackage[T1]{fontenc}    
\usepackage{hyperref}       
\usepackage{url}            
\usepackage{booktabs}       
\usepackage{amsfonts}       
\usepackage{nicefrac}       
\usepackage{microtype}      
\usepackage{lipsum}		
\usepackage{graphicx}
\usepackage{natbib}
\usepackage{doi}
\usepackage{amsfonts} 
 \usepackage{amsmath}

\title{Unified token representations for sequential decision models}

\author{
    Zhuojing Tian$^*$\\
    Intelligent Game and Decision Lab(IGDL)\\ 
    Beijing, China\\
    \texttt{tianzhuojing@foxmail.com} \\
    \And
    Yushu Chen \thanks{Equal contribution.}\\
    Tsinghua University \\ 
    Department of Computer Science and Technology\\
    Beijing, China\\
    \texttt{chenyushu@mail.tsinghua.edu.cn}
}

\renewcommand{\shorttitle}{\textit{arXiv} Template}

\begin{document}
\maketitle

\begin{abstract}
Transformers have demonstrated strong potential in offline reinforcement learning (RL) by modeling trajectories as sequences of return-to-go, states, and actions. However, existing approaches such as the Decision Transformer(DT) and its variants suffer from redundant tokenization and quadratic attention complexity, limiting their scalability in real-time or resource-constrained settings. To address this, we propose a Unified Token Representation (UTR) that merges return-to-go, state, and action into a single token, substantially reducing sequence length and model complexity. Theoretical analysis shows that UTR leads to a tighter Rademacher complexity bound, suggesting improved generalization. We further develop two variants: UDT and UDC, built upon transformer and gated CNN backbones, respectively. Both achieve comparable or superior performance to state-of-the-art methods with markedly lower computation. These findings demonstrate that UTR generalizes well across architectures and may provide an efficient foundation for scalable control in future large decision models.
\end{abstract}


\section{Introduction}
Transformers \citep{ashish2017attention} have become a foundational architecture across diverse domains, including natural language processing (NLP) \citep{brown2020language} and computer vision (CV) \citep{hatamizadeh2023global}, due to their strong capability of modeling long-range dependencies. This strength has motivated their adaptation to reinforcement learning (RL), where agent–environment interactions naturally form temporal sequences. In offline RL, the Decision Transformer (DT)\citep{chen2021decision} and its variants\citep{kim2024decision,wang2025long,zhengdecision} reformulate policy learning as conditional sequence modeling, treating trajectories as ordered triplets of return-to-go (RTG), states, and actions.

However, encoding RTG, state, and action as three separate tokens triples the sequence length ($L \rightarrow 3L$) and incurs quadratic attention complexity, making Transformer-based RL architectures computationally expensive and difficult to scale in real-time or resource-constrained environments. Moreover, RL trajectories are inherently governed by local Markovian dependencies\citep{kim2024decision}, where applying global self-attention uniformly across tokens introduces redundancy without proportional performance gains.

To address these limitations, we propose the Unified Token Representation (UTR), which fuses RTG, state, and action into a single compact token at each timestep. This unified encoding substantially reduces sequence length and model complexity while preserving expressiveness. From a theoretical perspective, we show that UTR yields a tighter Rademacher complexity bound, suggesting enhanced generalization in policy learning.

Building upon UTR, we develop two complementary variants: Unified Decision Transformer (UDT) and Unified Decision Conv (UDC), based on Transformer and gated convolutional backbones, respectively. UDT preserves the global modeling capacity of Transformers while leveraging unified token representations to shorten sequence length and reduce quadratic attention cost, thereby improving efficiency without compromising long-horizon reasoning. UDC further replaces global attention with a Gated Depthwise Convolutional Module that captures local temporal dependencies in linear time, offering a lightweight inductive bias for efficient decision-making.

Extensive experiments on standard offline RL benchmarks, including MuJoCo and AntMaze, demonstrate that both UDT and UDC achieve comparable or superior performance to state-of-the-art methods, while drastically reducing training and inference costs. These findings show that UTR generalizes effectively across architectures and may provide a scalable foundation for efficient large decision models.

In summary, our main contributions are threefold:
\begin{itemize}
    \item We propose UTR, a unified token representation that merges return-to-go, state, and action, substantially reducing sequence redundancy and computation.  
    \item We theoretically show that UTR achieves a smaller Rademacher complexity bound, indicating stronger generalization capacity.  
    \item We introduce two architectural variants: UDT(Transformer-based) and UDC(Gated CNN-based), empirically demonstrate their superior efficiency–performance trade-offs on offline RL benchmarks.
\end{itemize}

The remainder of this paper is organized as follows. Section 2 reviews related work. Section 3 introduces the methodology, including unified token representation and the Gated-CNN decision module. Section 4 presents the experimental setup, results, and discussion. Finally, Section 5 concludes the paper and outlines future research directions.

\section{Related Work}
\label{sec:related}
\subsection{Offline Reinforcement Learning with Transformer Variants}
Transformers have been effectively introduced into offline reinforcement learning by modeling trajectories as sequences of RTG, states, and actions~\citep{chen2021decision}. This formulation enables long-horizon credit assignment via temporal self-attention but introduces quadratic computational complexity and triples the sequence length due to token triplets $(R_t, s_t, a_t)$, limiting scalability and real-time applicability.  

Subsequent works have attempted to improve efficiency through linearized or kernelized attention mechanisms~\citep{wang2020linformer, choromanski2020rethinking, katharopoulos2020transformers}, or by combining attention with convolutional operators~\citep{kim2024decision, ota2024decision}. However, these approaches typically assume fixed sequence structures and uniform temporal operators, overlooking the redundancy among RTG, state, and action tokens.  

In contrast, our method introduces a unified token that fuses RTG, state, and action information, effectively reducing sequence redundancy and modality heterogeneity. Together with gated depthwise convolutions that replace self-attention, our model achieves efficient temporal reasoning and improved scalability while maintaining strong representational power.

\subsection{Gated CNNs for Conditional Sequence Modeling}
Gated convolutional architectures have emerged as efficient alternatives to attention-based models for sequence modeling. Early works such as WaveNet~\citep{van2016wavenet} and Gated CNNs for language modeling~\citep{dauphin2017language} demonstrated that multiplicative gating enables the capture of both local and long-range dependencies without explicit recurrence or attention. Recent developments, including ConvNeXt~\citep{liu2022convnet}, ConvNeXt V2~\citep{woo2023convnext}, and ModernTCN~\citep{luo2024moderntcn}, further highlight the potential of depthwise convolutions as efficient token mixers across modalities.  

In reinforcement learning, hybrid architectures such as Decision Convformer(DC)~\citep{kim2024decision} and Decision Mamba(DMamba)~\citep{ota2024decision} combine attention or state-space mechanisms with convolutions to capture both short- and long-term dependencies. While effective, these models incur significant parameter and memory overhead. Recent findings~\citep{yu2025mambaout} also show that for short causal sequences common in RL, gated CNNs achieve superior efficiency and comparable performance. 

Motivated by these insights, we propose a fully convolutional, RL-oriented architecture that integrates unified tokenization with gated depthwise convolutions, enabling adaptive fusion of multi-scale dependencies with significantly reduced computational cost and latency—making it well-suited for scalable offline and real-time decision-making.

\section{Methodology}
\label{sec: method}
\subsection{Preliminaries}

{\bf Offline Reinforcement Learning.} 
Offline Reinforcement Learning(Offline RL) aims to learn effective policies from a fixed dataset collected by one or more behavior policies, without further interactions with the environment \citep{prudencio2023survey}. Formally, an RL problem is modeled as a Markov Decision Process (MDP), defined by the tuple $(\mathcal{S}, \mathcal{A}, P, r, \gamma)$, where $\mathcal{S}$ is the state space, $\mathcal{A}$ is the action space, $P(s'|s, a)$ denotes the transition dynamics, $r(s, a)$ is the reward function, and $\gamma \in [0,1)$ is the discount factor. The agent's goal is to learn a policy $\pi(a|s)$ that maximizes the expected cumulative reward. In the offline setting, the algorithm only has access to a static dataset $\mathcal{D} = \{(s_t, a_t, r_t, s_{t+1})\}$, often collected under suboptimal or unknown policies. This setting introduces challenges such as distributional shift and extrapolation errors, which render classical online RL algorithms unstable or inapplicable.

{\bf Decision Transformer.} 
DT reinterprets policy learning as a sequence modeling problem, inspired by the success of Transformers in NLP. In this framework, trajectories are represented as ordered sequences of return-to-go (RTG), states, and actions, and the model autoregressively predicts future actions conditioned on these tokens. Specifically, the input to the model at timestep $t$ is the token triplet $(R_t, s_t, a_t)$, where $R_t = \sum_{t'=t}^{T} r_{t'}$ denotes the target return-to-go. Each token type is independently embedded and processed by a standard Transformer decoder, where self-attention enables the model to capture temporal dependencies across the trajectory. 

While DT has shown competitive performance on various offline RL benchmarks, its reliance on three separate tokens per timestep triples the sequence length, leading to quadratic complexity in self-attention computation. This design creates a bottleneck for long-horizon tasks and real-time deployment, motivating the development of more efficient tokenization and sequence-processing strategies.

{\bf Gated Depthwise Convolution.}
Convolutional networks have recently emerged as efficient alternatives to transformers for sequential modeling. Pioneering studies, such as Gated Convolutional Networks \citep{dauphin2017language}, demonstrated that multiplicative gating enables competitive sequence modeling without explicit attention. More recent architectures, including ModernTCN \citep{luo2024moderntcn}, ConvNeXt \citep{liu2022convnet}, and ConvNeXt V2 \citep{woo2023convnext}, have shown that properly designed convolutional backbones can rival transformer-based architectures in efficiency and performance across vision and time-series tasks.

In reinforcement learning, Gated Depthwise Convolution (Gated-CNN) modules combine channel-wise (depthwise) filtering with learned multiplicative gates that modulate information flow dynamically. Depthwise filters efficiently capture local temporal dependencies, while gating mechanisms emphasize salient features and suppress noise. This structure aligns naturally with the predominantly local, Markovian nature of RL trajectories, while stacking, dilation, or residual connections allow the capture of longer-range dependencies. Empirical evidence shows that lightweight local architectures, such as gated CNNs, can outperform more complex attention-based designs in both efficiency and generalization under short-context or real-time scenarios \citep{kim2024decision,yu2025mambaout}.

Collectively, these findings suggest that Gated-CNN modules can serve as a scalable and effective token-mixing primitive for RL, preserving decision performance while substantially reducing computational cost, latency, and memory footprint compared to full self-attention mechanisms. This motivates our proposed method, which integrates unified tokenization with Gated-CNN for efficient and scalable offline reinforcement learning.

\subsection{Unified Token Representation}
\label{sec: unified token}
We begin by encoding the scalar return-to-go $R_t$ into a low-dimensional vector representation:
\begin{equation}
    e^R_t = \sigma(\text{Linear}_R(R_t)),
\end{equation}
where $\text{Linear}_R(\cdot)$ projects the scalar return into a latent space (e.g., 32 dimensions), and $\sigma(\cdot)$ denotes a sigmoid gate. The projection matrix inherently contains coefficients of varying magnitudes, enabling dimensions associated with smaller weights to retain sensitivity even for large return inputs. Consequently, each latent dimension can contribute differentially to downstream prediction, ensuring that no component of the return signal is completely saturated after gating.

To align temporal dependencies for action prediction, we shift the action sequence one step forward:
\begin{equation}
    \tilde{a}_t = 
    \begin{cases}
    0, & t = 1, \\
    a_{t-1}, & t > 1,
    \end{cases}
\end{equation}
so that the model predicts the current action based on the current state and the corresponding return signal, consistent with the autoregressive decision formulation.

We then concatenate the gated return embedding, the current state, and the shifted action to form a unified representation:
\begin{equation}
    x_t = [\, e^R_t ,\, s_t ,\, \tilde{a}_t \,],
\end{equation}
which encapsulates both the current environmental context and prior behavioral information relevant to decision prediction. This concatenated feature is projected into the model’s hidden space via a fusion layer:
\begin{equation}
    z_t = \text{Linear}_F(x_t),
\end{equation}
ensuring dimensional consistency and feature alignment across different modalities.

To retain temporal awareness, we introduce a learnable timestep embedding $e^T_t = \text{Embedding}(t)$ and add it to the fused feature:
\begin{equation}
    h_t = z_t + e^T_t.
\end{equation}
Finally, a Layer Normalization operation standardizes the resulting token:
\begin{equation}
    \tilde{h}_t = \text{LayerNorm}(h_t),
\end{equation}
The module produces the unified token representation $\{\tilde{h}_1, \tilde{h}_2, \dots, \tilde{h}_L\}$ with shape $[B, L, D]$, 
where $B$ denotes the batch size, $L$ represents the sequence length, and $D$ corresponds to the feature embedding dimension of each token.

This unified formulation achieves three key effects. First, it restores the original sequence length of the trajectory, reducing self-attention complexity from $\mathcal{O}(9L^2)$ to $\mathcal{O}(L^2)$, and proportionally lowering the computational cost of convolution-based mixers. Second, the gated return embedding adaptively modulates the influence of reward expectations while preserving gradient sensitivity across scales. Third, the shifted-action design ensures correct causal alignment for autoregressive action prediction. 

\begin{figure*}[t]
\begin{minipage}[b]{\linewidth}
	\centering
        \includegraphics[width=0.8\linewidth]{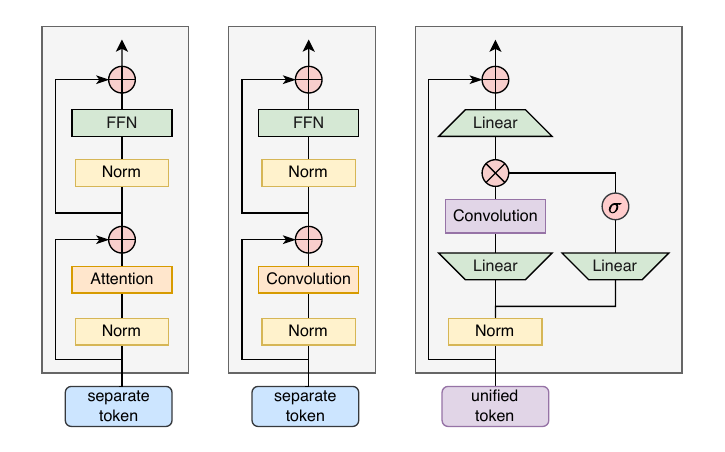}
	    \caption{$\textbf{\emph{Left:}}$ Decision Transformer. $\textbf{\emph{Middle:}}$ Decision Convformer. $\textbf{\emph{Right:}}$ Gated CNN Decision Module.}
	\label{fig:fig2}
\end{minipage}
\end{figure*}

\subsection{Gated CNN Decision Module}
\label{sec: gated cnn}
As illustrated in Figure~\ref{fig:fig2}, the left, middle, and right panels depict the architectures of DT, DC, and our proposed Gated CNN model, respectively. Both DT and DC adopt the MetaFormer framework~\citep{yu2022metaformer}, where each block consists of normalization, a token mixer, and a feed-forward network. DT employs quadratic-cost self-attention, whereas DC replaces attention with static causal convolutions to improve efficiency, but both rely on rigid or computationally intensive token interaction mechanisms, limiting adaptability in short-horizon decision tasks.

Recent work MambaOut~\citep{yu2025mambaout} shows that for short causal sequences ($L \ll 6D$), gated convolutional architectures achieve higher modeling efficiency and competitive dependency extraction compared with state-space or attention-based mechanisms. Motivated by this, we construct a pure Gated CNN architecture, where temporal dependencies are captured directly through gated convolutions, achieving linear-time computation while preserving adaptability to dynamic decision contexts.

Formally, let $X \in \mathbb{R}^{L \times D}$ denote a sequence of $L$ unified tokens with embedding dimension $D$; after layer normalization $\hat{X} = \mathrm{LN}(X)$, the normalized sequence is projected into two parts, one sent to a causal depthwise separable convolution $H = \mathrm{DWConv}(\hat{X}; K)$, and the other used for the gating branch $G = \mathrm{SiLU}(\hat{X})$, then combined with the input via a residual connection $Y = W_o (H \odot G) + X$, where $\odot$ denotes element-wise multiplication and $W_o$ is the output projection, thus preserving temporal causality while introducing unified token representation and causal depthwise separable convolution for efficient modeling.

Compared with the Mamba block~\citep{ota2024decision}, which integrates both state-space modeling (SSM) and gating mechanisms, our Gated CNN block omits the SSM component and focuses purely on gated convolutional dynamics, yielding lower parameter count and computational overhead while retaining strong locality modeling capabilities essential for short decision horizons. In summary, the proposed Gated CNN Decision Module provides a lightweight yet expressive alternative to attention- or SSM-based architectures, achieving an effective balance between modeling capacity, computational efficiency, and adaptability, making it particularly suitable for real-time or resource-constrained decision-making environments.


\subsection{Theoretical Analysis}
A central question in representation learning is whether different tokenization strategies affect the ability of the model to generalize from limited data.  
Rademacher complexity provides a principled measure of the capacity of a hypothesis class: a lower Rademacher complexity generally implies a tighter generalization bound.  
In this section, we establish a simplified but representative assumption under which we can prove that the merged-token representation exhibits strictly lower Rademacher upper bound than the separated-token representation.  

\paragraph{Rademacher complexity.}
Let \(\mathcal{F}\) be a class of functions mapping from an input space \(\mathcal{X}\) to \(\mathbb{R}\), and let \(S=\{x_1,\ldots,x_n\}\) be a sample of size \(n\) drawn i.i.d. from some distribution over \(\mathcal{X}\).  
The empirical Rademacher complexity of \(\mathcal{F}\) with respect to \(S\) is defined as
\begin{equation}
\widehat{\mathcal{R}}_S(\mathcal{F})
= \mathbb{E}_\sigma\Bigg[\,\sup_{f\in\mathcal{F}}\;\frac{1}{n}\sum_{i=1}^n \sigma_i f(x_i)\Bigg],
\end{equation}
where \(\sigma_1,\ldots,\sigma_n\) are independent Rademacher random variables taking values in \(\{\pm 1\}\) with equal probability.  
The expected Rademacher complexity is the expectation of \(\widehat{\mathcal{R}}_S(\mathcal{F})\) over the random sample \(S\).

\paragraph{Connection to generalization.}
Rademacher complexity directly controls the generalization gap between empirical risk and expected risk.  
A generalization bound is given in the following standard result.

\textbf{Theorem 1}(see, e.g., Shai and Shai, 2014): Let \(\mathcal{F}\) be a class of functions mapping \(\mathcal{X}\) to \([0,1]\).  
For any \(\delta>0\), with probability at least \(1-\delta\) over a sample \(S\) of size \(n\), the following holds for all \(f\in\mathcal{F}\):
\begin{equation}
    \mathbb{E}[f(x)] \;\le\; \frac{1}{n}\sum_{i=1}^n f(x_i)
+ 2\,\widehat{\mathcal{R}}_S(\mathcal{F})
+ 4\sqrt{\frac{2\log(4/\delta)}{n}}.
\end{equation}

This theorem implies that, if the Rademacher complexity of the unified representation class is lower than that of the separated representation class, then, under comparable capacity constraints, the unified representation enjoys a provably tighter generalization guarantee.  
Hence, our subsequent analysis of covariance structure and trace bounds provides not only a theoretical justification, but also a practical explanation for the empirical advantages of unified tokenization.

By establishing a simplified but representative assumption, it can be proven that the merged-token representation has strictly lower Rademacher upper bound than its separated-token counterpart.

Let \(\mathcal{F}\) denote the class of linear predictors on the input with a weight norm constraint \(\|v\|_2\le B\).  A standard Rademacher upper bound for linear classes (e.g. Bartlett and Mendelson, Mohri et al.) yields, up to universal constants,
\begin{equation}
    \mathcal{R}_n(\mathcal{F}) \le \mathcal{\hat{R}}_n(\mathcal{F}) = B \sqrt{\frac{\operatorname{Tr}(\operatorname{Cov}(\text{input}))}{n}},
\end{equation}
where \(n\) is the number of i.i.d.\ samples. In the following we compare these upper bounds for the two tokenizations.

\textbf{Theorem 2}: 
Let $z=\sum_{i=1}^3 w_i u^{(i)}$ denote the unified representation
and $\hat{z}=[u^{(1)};u^{(2)};u^{(3)}]$ the concatenated representation,
where each $u^{(i)}$ has covariance block $\Sigma_{ii}$ with
$\operatorname{Tr}(\Sigma_{ii}) \le T$ and pairwise correlation at most $\rho$,
and where the weight vector $w=(w_1,w_2,w_3)$ satisfies $\|w\|_2^2 = s$. 
The sample size $n$ and the weight-norm budget $B$ (or regularization policy) are identical when comparing the two tokenizations.
Then their covariance traces satisfy
\begin{align}
\operatorname{Tr}\!\big(\operatorname{Cov}(z)\big)
&\le T\big(\rho + (1-\rho)\,s\big), \label{eq:merged-trace} \\
\operatorname{Tr}\!\big(\operatorname{Cov}(\hat{z})\big)
&\le 3T. \label{eq:sep-trace}
\end{align}
Consequently, the Rademacher upper bounds satisfy
\begin{equation}
\frac{ \hat{\mathcal{R}}_n(\mathcal{F}_{\mathrm{merged}}) }
     { \hat{\mathcal{R}}_n(\mathcal{F}_{\mathrm{sep}}) }
\;\le\;
\sqrt{\frac{\rho + (1-\rho)\,s}{3}}.
\end{equation}

\textbf{Proof}: Compute the covariance of the merged vector:
\begin{equation}
\operatorname{Cov}(z)
= \operatorname{Cov}\Big(\sum_i w_i u^{(i)}\Big)
= \sum_{i,j} w_i w_j \Sigma_{ij},
\end{equation}
hence
\begin{equation}
\operatorname{Tr}\big(\operatorname{Cov}(z)\big)
= \sum_{i,j} w_i w_j \operatorname{Tr}(\Sigma_{ij}).
\end{equation}
Using \(\operatorname{Tr}(\Sigma_{ii})=T\) and \(\operatorname{Tr}(\Sigma_{ij})\le \rho T\) for \(i\ne j\),
\begin{align*}
\operatorname{Tr}\big(\operatorname{Cov}(z)\big)
&\le \sum_i w_i^2 T + \sum_{i\ne j} w_i w_j \rho T \\
&= T\Big(s + \rho\big(1-s\big)\Big)
= T\big(\rho + (1-\rho)s\big),
\end{align*}
where \(\sum_{i\ne j} w_i w_j = (\sum_i w_i)^2 - \sum_i w_i^2 = 1-s\). This proves \eqref{eq:merged-trace}.

For the concatenated vector \(X\),
\begin{equation}
\operatorname{Tr}\big(\operatorname{Cov}(X)\big)
= \sum_i \operatorname{Tr}(\Sigma_{ii})
\le 3T,
\end{equation}
which proves \eqref{eq:sep-trace}.

Combining these trace bounds with the standard linear Rademacher upper bound (which scales with the square root of the trace divided by \(n\)) yields the stated inequality for the ratio of the upper bounds.

\textbf{Remark}: The inequality above compares the theoretical Rademacher upper bounds for the two tokenizations (i.e. the right-hand side bounds the ratio of the bounds). It does not assert equality of the true Rademacher complexities. 

\paragraph{Discussion of assumptions and their plausibility.}
As noted above, although simplified this modelling assumption captures typical statistical patterns observed in learned embeddings and therefore provides a useful first-order explanation for the empirical benefits of the merged tokenization.  Below we briefly justify the three modelling choices used in the analysis.

\begin{itemize}
    \item \textbf{Linear predictor / linearized bound.}  
    We base the comparison on the standard linear-class Rademacher upper bound
    \(\widehat{\mathcal{R}}_n(\mathcal{F}) \propto B\sqrt{\operatorname{Tr}(\mathrm{Cov})/n}\),
    where an increase in the input covariance trace monotonically increases the bound.  
    For nonlinear predictors (e.g., transformer decoders) that are \(L\)-Lipschitz, Talagrand's contraction lemma (Talagrand, 1994) implies that the Rademacher complexity of the composed class is at most \(L\) times that of the linear class.  
    Thus, while exact equality does not hold, the linear-class bound provides a principled proxy for comparing tokenizations.

  \item \textbf{Unified representation as a weighted linear combination.}  
  Modelling the merged token as \(z=\sum_i w_i u^{(i)}\) is a compact abstraction of the common engineering
  implementation where one concatenates sub-embeddings and applies a linear projection (or block-wise
  reweighting). The scalar \(s=\sum_i w_i^2\) quantifies weight concentration;
  smaller \(s\) (more uniform weights) tightens the trace bound.

  \item \textbf{Approximate equality of diagonal traces.}  
  We assume \(\operatorname{Tr}(\Sigma_{ii})\approx T\) for notational simplicity. In practice, per-token
  normalization layers (e.g. LayerNorm) and standard preprocessing tend to make the per-type variances
  comparable, so the equal-trace approximation is reasonable. 
\end{itemize}




\section{Experiments}
\subsection{Experimental Setup}
We evaluate two model variants derived from the methods described in Section~\ref{sec: method}: 
\begin{itemize}
    \item \textbf{Decision Unified Transformer (DUT):} a Decision Transformer-style model that adopts unified token encoding without altering the standard Transformer architecture, as described in Section~\ref{sec: unified token}.
    \item \textbf{Decision Unified Conv (DUC):} extends DC by replacing the metaformer with a gated CNN architecture using causal depthwise separable convolutions, as detailed in Section~\ref{sec: gated cnn}, while maintaining the same unified tokenization scheme.
\end{itemize}

We evaluate our models on a diverse set of tasks from the D4RL benchmark~\cite{fu2020d4rl}, covering both continuous-control and sparse-reward domains:  
\begin{itemize}
    \item \textbf{MuJoCo Locomotion:} Hopper, HalfCheetah, Walker2d, and Ant under the \emph{medium}, \emph{medium-replay}, \emph{medium-expert}, and \emph{expert} settings.
    \item \textbf{AntMaze Navigation:} \emph{umaze} and \emph{umaze-diverse} configurations to assess generalization in sparse-reward environments.
\end{itemize}

These two models are evaluated against strong offline RL baselines, including DT, DC, and Decision Mamba (DMamba)~\cite{ota2024decision}.  
DT serves as the foundational return-conditioned Transformer model, DC introduces convolutional token mixing for improved efficiency, and DMamba represents a recent state-space variant that integrates selective gating mechanisms. This comparison allows us to assess the effectiveness of unified token encoding and gated CNN modeling under a consistent experimental framework. For all algorithms, we report normalized D4RL scores, where a score of 100 corresponds to expert-level performance.  
Following the evaluation protocol established in DC, the initial Return-to-Go (RTG) value during testing is treated as a tunable hyperparameter.  
Six target RTG values are examined—each being an integer multiple of the default RTG defined by Chen \textit{et al.}~\cite{chen2021decision}—and the highest normalized score among them is reported for each algorithm. Additional details regarding hyperparameter configurations, model sizes, and training settings are provided in the Appendix.

\subsection{Results and Analysis}

\begin{table*}[h]
\centering
\begin{tabular*}{\textwidth}{@{\extracolsep{\fill}}cccccc}
\toprule
\textbf{Dataset} & \textbf{DT}  & \textbf{DC} & \textbf{DMamba} & \textbf{DUT*} & \textbf{DUC*}\\ 
\cmidrule(lr){1-6}
\textbf{HalfCheetah-m}    & 42.6 & \underline{42.9} & 42.8 & 42.9 & \textbf{43}  \\                      
\textbf{Hopper-m}         & 67.6 & \textbf{94.5} & 83.5 & 79.4 & \underline{86.5}  \\                 
\textbf{Walker-m}         & 74 & \textbf{79.5} & 78.2 & 77.1 & \underline{78.2}   \\ \midrule              
\textbf{HalfCheetah-m-r}  & 36.6 & \underline{41.3} & 39.6 & 38.9 & \textbf{41.7} \\
\textbf{Hopper-m-r}       & 82.7 & 85 & 82.6 & \textbf{94.2} & \underline{85.8}   \\              
\textbf{Walker-m-r}       & 66.6 & 75 & 70.9 & \underline{76.4} & \textbf{76.9}   \\ \midrule
\textbf{HalfCheetah-m-e}  & 86.8 & 89 & 91.9 & \underline{91.9} & \textbf{92.8}  \\                   
\textbf{Hopper-m-e}       & 107.6 & 109.4 & \underline{111} & 109.2 & \textbf{111} \\                     
\textbf{Walker-m-e}       & 108.1 & \textbf{109.1} & 108.3 & \underline{110.5} & 107.8  \\ \midrule                    
\textbf{ant-e}            & 123.1 & 126.5 & \textbf{130} & \underline{127.6} & 126.7   \\                    
\textbf{ant-m}            & 95.3 & \underline{96.4} & 86.1 & \textbf{96.6} & 95      \\                 
\textbf{ant-m-e}          & \underline{129.3} & 127.6 & 129 & 126 & \textbf{129.4}   \\                 
\textbf{ant-m-r}          & 81.4 & \textbf{97} & 88.3 & 92.4 & \underline{93.4}   \\   \midrule                 
\textbf{antmaze-umaze}    & 69.8 & 76 &\underline{79} & 71 & \textbf{80}  \\                       
\textbf{antmaze-umaze-d}  & 70.3 & 66 & \textbf{80} & 65 & \underline{78}  \\  \bottomrule                    
\end{tabular*}
\caption{Overall Performance. m, e, m-r, and m-e denote the medium, expert, medium-replay, and medium-expert; u and u-d denote the umazed and umazed-diverse, respectively. Methods marked with * are designed by us, bold and underline indicate the highest score and the second-highest score.}
\label{table1}
\end{table*}

\textbf{MuJoCo locomotion benchmarks:} 
As shown in Table~\ref{table1}, compared to DT, DUT consistently improves performance across most locomotion tasks, particularly on \textit{Hopper-medium} and \textit{Walker2d-medium-replay}, demonstrating the effectiveness of unified token encoding in reducing sequence length while preserving trajectory consistency. Building on this, DUC further surpasses DUT, DC, and DMamba across the majority of MuJoCo datasets. The gains are most pronounced in \textit{Hopper} and \textit{Walker2d} series, highlighting the benefit of the gated CNN in modeling short-horizon causal dependencies. On \textit{Ant} tasks, where the dynamics are more complex and actions are high-dimensional, DUC achieves performance comparable to DMamba but with lower computational cost, indicating better efficiency–accuracy trade-offs. Overall, these results confirm that combining unified tokenization with depthwise separable gated convolutions enhances both representational efficiency and generalization in continuous-control offline RL.

\textbf{AntMaze:} 
For the \textit{AntMaze} tasks, which involve long-horizon navigation under sparse rewards, DMamba achieves the best score on \textit{umaze-diverse} due to its structured state-space modeling for temporal abstraction. Nevertheless, DUC attains competitive results across both \textit{umaze} and \textit{umaze-diverse}, maintaining a much simpler architecture and lower computational overhead. This confirms that unified tokenization and depthwise separable gated convolutions provide strong generalization and stability even in challenging sparse-reward, long-horizon offline RL settings.

\subsection{Efficiency Analysis}
Table~\ref{table2} presents a comparative analysis of computational efficiency among DT, DUT, DC, and DUC on \texttt{hopper-medium}, evaluated on a single NVIDIA RTX A6000 GPU. The reported time corresponds to the 500-step training duration. Compared to DT, DUT reduces FLOPs by 67.34\% but achieves only a modest 5.56\% reduction in time through unified tokenization. This discrepancy arises because modern GPUs exhibit strong parallel processing capabilities, and additional factors such as I/O latency and memory bandwidth limitations can further mask theoretical computational savings when the model size is relatively small. Nonetheless, as model scale increases, the impact of parallelism diminishes and the time efficiency gains from reduced computational complexity are expected to become more pronounced. DUC further achieves a 74.92\% FLOP reduction and a 30.02\% speedup over DC by combining unified tokenization with gated depthwise convolutions, which replace global attention with localized linear-time operations. This design not only preserves modeling capacity but also substantially lowers computational and memory costs. Overall, these results highlight the superior scalability of the unified token–based convolutional architecture, making it particularly suitable for large-scale or resource-constrained offline RL applications.

\begin{table*}[h]
\centering
\begin{tabular*}{\textwidth}{@{\extracolsep{\fill}}ccccccc}
\toprule
\textbf{Complexity} & \textbf{DT} & \textbf{DUT} & \textbf{\textbf{$\triangle\%$}} & \textbf{DC} & \textbf{DUC} & \textbf{$\triangle\%$} \\
\cmidrule(lr){1-7}  
Time(s)         & 6.12 & 5.78 & 5.56\%  & 4.93 & 3.45 & 30.02\% \\
FLOPs(Billion)  & 9.46 & 3.09 & 67.34\% & 6.14 & 1.54 & 74.92\% \\
params(million) & 2.63 & 2.63 & 0.00\%  & 1.99 & 1.46 & 26.63\% \\ \bottomrule  
\end{tabular*} 
\small\caption{Comparison of time, FLOPs, and parameters for DT, DC, DUT, DUC on Hopper-m.}
\label{table2}
\end{table*}

\section{Conclusion and Future Work}
\label{sec:conclusion}
This paper presents two complementary components for efficient offline reinforcement learning: a Unified Token Representation that jointly encodes return-to-go, state, and action information to reduce sequence redundancy, and a Gated CNN Decision Module that leverages lightweight convolution to capture temporal dependencies effectively. Together, these components enable compact, scalable, and resource-efficient policy learning while maintaining strong performance across D4RL benchmarks. Notably, the Unified Token Representation also has the potential to serve as a foundational building block for future large-scale decision-making models. In future work, we plan to explore extending UTR to larger models and more complex environments, as well as integrating it with advanced modeling techniques to further improve scalability and generalization in high-dimensional decision tasks.



\bibliographystyle{unsrtnat}
\bibliography{references}  

\end{document}